\documentclass[11pt, letterpaper]{article}
\usepackage[utf8]{inputenc}
\usepackage[margin=1.5cm]{geometry}
\usepackage{titlesec}
\usepackage{tabu}
\usepackage{enumitem}
\usepackage{booktabs, multirow, makecell}
\usepackage{amsmath}
\usepackage{amssymb}
\usepackage{xcolor}
\newlist{selectlist}{itemize}{2}
\setlist[selectlist]{label=$\square$,leftmargin=*,noitemsep,topsep=0pt}

\usepackage{lmodern}
\usepackage{multirow}
\usepackage{graphicx}
\usepackage{hyperref}
\hypersetup{
    colorlinks=true,
    linkcolor=blue,
    filecolor=magenta,      
    urlcolor=blue,
}
 
\titleformat{\section}[block]{\hspace{1em}\bfseries}{\thesection.}{0.5em}{} 
\titleformat{\subsection}[block]{\hspace{1em}}{\thesubsection}{0.5em}{}

\begin{document}

\noindent
\textbf{\textit{Title / name of your software}}\\
\textbf{UnBias-Plus}: Detect, Explain, and Rewrite Bias

\vskip0.5cm
\noindent

\textbf{\textit{Names of authors / main developers (incl. affiliations, addresses, email)}}\\
Ahmed Y. Radwan$^{1}$, Ahmed ElKady$^{1}$, Sindhuja Chaduvula$^{2}$, Mohamed Hafez$^{1}$, Amrit Krishnan$^{1}$,\\ Shaina Raza$^{1,*}$\\
\{ahmed.radwan, ahmed.elkady, mohamed.hafez, amritk, shaina.raza\}@vectorinstitute.ai\\
\noindent
$^{1}$Vector Institute for Artificial Intelligence, Toronto, Canada; \quad 
$^{2}$Independent Researcher\\
$^{*}$Corresponding author

\noindent
\textbf{Abstract}\\
Bias in natural language remains a persistent challenge in both human-written and AI-generated content, affecting domains such as journalism, education, and AI research. Most existing detection methods identify only the presence of bias, with limited support for granular detection, interpretable explanations, neutral rewriting, and openly available trained models. We present\textbf{ UnBias-Plus}, an open-source toolkit unifying (1) segment-level multi-class bias classification, (2) biased span localization, (3) neutral text rewriting, and (4) reasoning for each decision. Available via Python, CLI, REST API, and web interfaces, UnBias-Plus supports accessible bias analysis. The toolkit, source code, models, datasets, and documentation are publicly available.
\vskip0.5cm


\vskip0.5cm
\noindent
\textbf{Code metadata}\\
\noindent
\begin{tabular}{|l|p{6.5cm}|p{9.5cm}|}
\hline
\textbf{Nr.} & \textbf{Code metadata description} & \textbf{Please fill in this column} \tabularnewline
\hline
C1 & Current code version & v0.1.6 \tabularnewline
\hline
C2 & Permanent link to code/repository used for this code version & \url{https://github.com/VectorInstitute/unbias-plus/tree/626f908} \tabularnewline
\hline
C3 & Permanent link to Reproducible Capsule & --- \tabularnewline
\hline
C4 & Legal Code License & Vector Institute License. The software may be accessed and used by Academic Entities, Sponsors, and Partners of the Vector Institute under the terms provided in \texttt{LICENSE.md} in the repository. \tabularnewline
\hline
C5 & Code versioning system used & git \tabularnewline
\hline
C6 & Software code languages, tools, and services used & Python, JavaScript, GitHub, Hugging Face, FastAPI \tabularnewline
\hline
C7 & Compilation requirements, operating environments \& dependencies & Python $\geq$3.10, $<$3.12; GPU with CUDA 12.4 recommended for model inference and fine-tuning; CPU execution supported but slower \tabularnewline
\hline
C8 & If available Link to developer documentation/manual & Documentation: \url{https://vectorinstitute.github.io/unbias-plus/}. Release version: \url{https://github.com/VectorInstitute/unbias-plus/releases/tag/v0.1.6} \tabularnewline
\hline
C9 & Support email for questions & shaina.raza@vectorinstitute.ai \tabularnewline
\hline
\end{tabular}\\
\vskip0.5cm
\noindent

\section{Introduction}

Bias in natural language can influence how people and groups are represented, interpreted, and treated across human-written and AI-generated content~\cite{gallegos2024bias}. Earlier text-bias research commonly framed the problem as binary or multi-class classification tasks, including sentence-level bias detection, span-level identification, and neutral rewriting from annotated examples~\cite{gallegos2024bias}. Generative AI broadens this challenge because outputs are open-ended, context-dependent, and shaped by the prompt and intended use. Commercial moderation APIs~\cite{openai_moderation_api} and LLM guardrails~\cite{inan2023llama} can flag content against predefined safety policies or risk taxonomies, but are generally designed for broad safety enforcement rather than detailed bias analysis. Moreover, many widely used moderation solutions are proprietary, while open alternatives often focus on safety classification rather than bias characterization and remediation. There remains a need for open, reusable software that identifies bias at a granular, span level.

We present \textbf{UnBias-Plus} (UnBias$^{+}$), an open-source Python toolkit for fine-grained bias analysis and rewriting. The toolkit provides fine-tuned models trained on a purpose-built annotated dataset and supports reproducible end-to-end workflows for bias detection, explanation, and neutral rewriting. It supports segment-level bias localization, multi-class bias-type classification, severity estimation, reasoning for each detected segment, and neutral rewriting of biased spans. The toolkit provides structured outputs through a Python package, command-line interface, REST API, and web interface, enabling integration into research workflows and text-processing applications. The current release supports fine-tuned models from the Qwen3 family~\cite{yang2025qwen3}, including Qwen3-8B and Qwen3.5-4B~\cite{qwen3.5}, and is adaptable to any instruction-tuned model.

\begin{figure}[t]
  \centering
  \includegraphics[width=\linewidth]{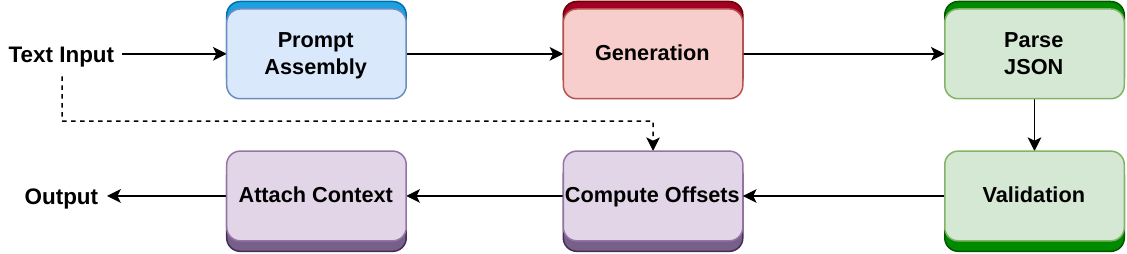}
  \caption{Pipeline architecture of UnBias-Plus. Input text flows
    through prompt construction, LLM inference, JSON parsing,
    offset computation, and result assembly.}
  \label{fig:pipeline}
\end{figure}

\begin{figure}[h]
  \centering
  \includegraphics[width=0.95\textwidth]{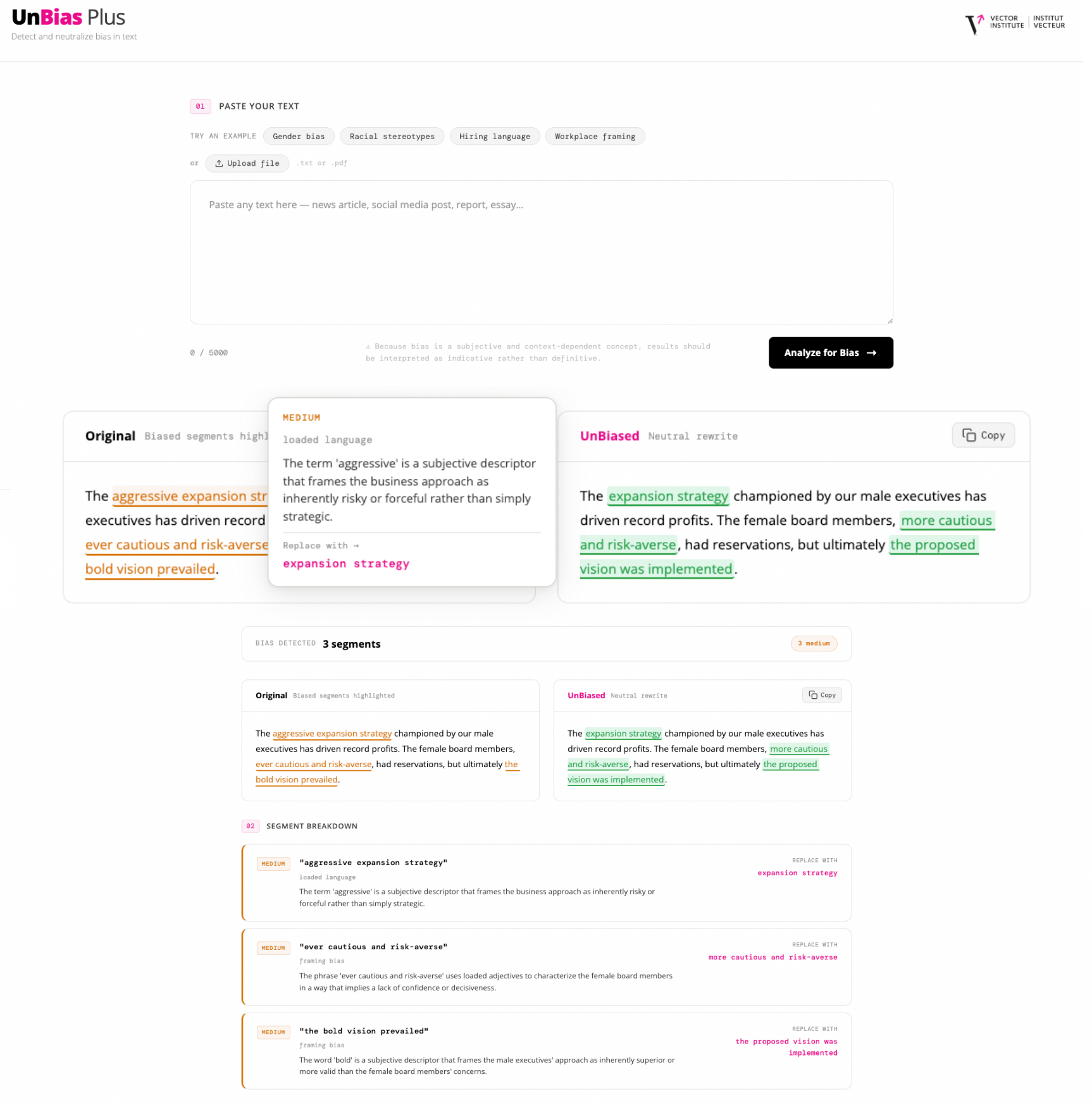}
  \caption{The web demo interface of UnBias-Plus allows users to input
    text, highlights biased segments with severity colour-coding,
    displays per-segment bias types and reasoning, and generates a full
    neutral rewrite.}
  \label{fig:ui_demo}
\end{figure}

\section{Software description}

UnBias-Plus implements a multi-stage bias review pipeline (Figure~\ref{fig:pipeline}). Given input text, the system prompts a fine-tuned language model to produce a structured analysis. The analysis includes whether bias is present, relevant spans with character-level offsets, bias types, severity scores, explanations, and suggested neutral replacements. For each detected segment, the output includes the original phrase, bias type, severity, explanation, and replacement, alongside a full neutral rewrite for holistic comparison. Results can be displayed in the web interface (Figure~\ref{fig:ui_demo}), printed via the command line, or exported as JSON for annotation, evaluation, dataset construction, and downstream model retraining.

\section{Example usage and validation}

Usage examples for the command-line interface, Python API, REST API, and web interface are provided in the documentation. To validate our model, we evaluate the released models on text outside the training distribution and whether the outputs support the intended bias review workflow.
We evaluated the released models on an out of distribution test set from BABE \cite{spinde-etal-2021-neural-media}, which is news media dataset. We took a stratified sample containing 175 biased and 175 unbiased samples. GPT-4o-mini was used as a judge on a 0--5 scale, alongside automatic metrics including recall at biased words, ROUGE-L, length ratio, hallucination rate, and duplicate rate. The prompts for the judge metrics are given in our codebase.

\begin{table}[t]
\centering
\caption{Unbias-Plus evaluation. $\uparrow$\,=\,higher is better, $\downarrow$\,=\,lower is better, ${\approx}1.0$\,=\,closest to 1.0 is best.}
\label{tab:ood-validation}
\resizebox{\textwidth}{!}{
\begin{tabular}{@{}l l cccccc cccc cccc@{}}
\toprule
& & \multicolumn{6}{c}{\textbf{Biased Samples}} & \multicolumn{4}{c}{\textbf{Unbiased Samples}} & \multicolumn{4}{c}{\textbf{Segment Level}} \\
\cmidrule(lr){3-8} \cmidrule(lr){9-12} \cmidrule(lr){13-16}
& \textbf{Model}
& \makecell{Bias\\red.\,\%\,$\uparrow$} & \makecell{Bias red.\\mean\,$\uparrow$} & \makecell{Relev.\,$\uparrow$} & \makecell{Global\\rew.\,$\uparrow$} & \makecell{ROUGE\\-L\,$\uparrow$} & \makecell{Len.\\ratio\,${\approx}1$}
& \makecell{Corr.~ID\\mean\,$\uparrow$} & \makecell{Corr.~ID\\med.\,$\uparrow$} & \makecell{Unnec.\\rew.\,$\uparrow$} & \makecell{Unnec.\\med.\,$\uparrow$}
& \makecell{Recall\\words\,$\uparrow$} & \makecell{Seg.\\qual.\,$\uparrow$} & \makecell{Halluc.\\rate\,$\downarrow$} & \makecell{Dupl.\\rate\,$\downarrow$} \\
\midrule
& Qwen3.5-4B$_{\text{UnBias}^{+}}$ & 57.5 & 1.91 & 4.09 & \textbf{3.15} & 0.71 & \textbf{1.00} & 3.44 & 2.0 & 3.97 & 3.0 & \textbf{0.85} & \textbf{3.88} & 5.3 & \textbf{0.0} \\[4pt]
& Qwen3-8B$_{\text{UnBias}^{+}}$ & \textbf{60.6} & \textbf{1.94} & \textbf{4.18} & 3.08 & \textbf{0.72} & 0.99 & \textbf{4.16} & \textbf{5.0} & \textbf{4.44} & \textbf{5.0} & 0.71 & 3.58 & \textbf{3.6} & 0.7 \\
\bottomrule
\end{tabular}
}
\vspace{2pt}
\par\raggedright\footnotesize
\textit{Metrics:} Bias red.\,\%/mean = bias removed (frac./raw); Relev.\ = semantic fidelity (1--5); Global rew.\ = rewrite quality (1--5); ROUGE-L = lexical overlap (0--1); Len.\ ratio = rewrite/orig.\ length; Corr.~ID = correct unbiased detection (1--5); Unnec.~rew.\ = avoids needless edits (1--5); Recall words = biased-word coverage (0--1); Seg.~qual.\ = segment edit quality (1--5); Halluc./Dupl.\ = fabricated/repeated content (\%).
\end{table}

Results in Table~\ref{tab:ood-validation} show a trade-off between bias detection/preservation and segment-level editing precision. Our fine-tuned Qwen3-8B UnBias-Plus achieves stronger bias reduction, higher relevance, stronger preservation of unbiased samples, and lower hallucination rate. The fine-tuned Qwen3.5-4B UnBias-Plus, however, improves rewrite and segment-level behaviour, with higher global rewrite quality, higher recall at biased words, stronger segment replacement quality, and zero duplicate rate. These results support the design of UnBias-Plus as a configurable toolkit: users can select a model variant depending on whether their workflow prioritizes conservative bias review, unbiased-text preservation, or more aggressive segment-level localization.

\section{Impact}
UnBias-Plus supports both technical and non-technical users through a unified workflow for bias detection, explanation, and neutral rewriting. It can be accessed through programmatic interfaces and an interactive web application, enabling integration into research, development, and practitioner-facing review processes.
The impact of UnBias-Plus spans research, industry, and society. From a research perspective, it provides a unified framework for bias detection, classification, explanation, and debiasing, together with associated models, code, and datasets that support reproducible development of fairness-aware language systems. In industry, it can be applied to domains such as journalism, content moderation, and decision-support systems to identify potentially biased language and improve transparency. More broadly, UnBias-Plus can promote equitable information dissemination and support fairer communication by reducing the propagation of biased language in public-facing systems.

\section{Limitations and Future Development}

UnBias-Plus should be used as a bias review assistant rather than as a final authority on neutrality or fairness. Bias is context-dependent: the same phrase may be interpreted differently across domains, communities, and cultural settings. For this reason, the toolkit is designed to support human-in-the-loop review, where model outputs can be inspected, contested, and revised before use in editorial, research, or decision-support workflows.
From a\textbf{ practical }perspective,  the amount of context that can be processed depends on the selected model and available hardware. While the released 4B and 8B variants make the toolkit easier to deploy, smaller models may still miss subtle framing patterns or produce incomplete reasoning for complex articles. Domain sensitivity is another limitation. In order to prepare it for a custom domain, it needs retraining the models, for which we have provided the code already.

Future development will focus on expanding domain coverage, improving multilingual support, and integrating additional model backends like multimodals~\cite{raza2025humanibench,radwan2026sonico1} . We also plan to extend the broader UnBias-Plus framework beyond text toward image and multimodal analysis, where bias, framing, and manipulation can appear across visual as well as linguistic content. 
To support this extensibility, the project releases the dataset, software pipeline, and training recipes so that users can adapt the toolkit to new models, domains, and deployment settings.

\section{Acknowledgment}

Resources were provided in part by the Province of Ontario and the Government of Canada through CIFAR and Vector Institute sponsors (\url{http://www.vectorinstitute.ai/\#partners}), and by the European Union Horizon Europe programme under the AIXPERT project (Grant No.\ 101214389).

We thank Orli Namian, Natalie Richard, Marcie De Cesare, Kylie Williams and Aravind Narayanan for their valuable support to the UnBias-Plus project.

\bibliographystyle{plain}
\bibliography{references}

\end{document}